\documentclass{ieeeaccess}
\usepackage{cite}
\usepackage{amsmath,amssymb,amsfonts}
\usepackage{graphicx}
\usepackage[unicode=true,colorlinks=true,linkcolor=blue,urlcolor=blue,citecolor=blue]{hyperref}
\usepackage{cleveref}
\usepackage[ruled,vlined,linesnumbered]{algorithm2e}
\usepackage{booktabs}
\usepackage{multirow}
\usepackage{multicol}
\usepackage{subcaption}
\usepackage[inline]{enumitem}
\usepackage{mathpartir}
\usepackage{makecell}
\usepackage{bm}
\usepackage{xspace}
\usepackage{pifont}
\usepackage{nicefrac}
\usepackage{rotating}
\usepackage{etoolbox}
\makeatletter
\patchcmd{\ps@titlepage}{VOLUME\ \thevol, 2016}{Accepted manuscript}{}{\PackageError{LLM-Flax}{Review footer patch failed}{}}
\patchcmd{\ps@titlepage}{VOLUME\ \thevol, 2016}{Accepted manuscript}{}{\PackageError{LLM-Flax}{Review footer patch failed}{}}
\patchcmd{\ps@headings}{VOLUME\ \thevol, 2016}{Accepted manuscript}{}{\PackageError{LLM-Flax}{Review footer patch failed}{}}
\patchcmd{\ps@headings}{VOLUME\ \thevol, 2016}{Accepted manuscript}{}{\PackageError{LLM-Flax}{Review footer patch failed}{}}
\patchcmd{\@maketitle}{Digital Object Identifier\space\@doi}{}{}{\PackageError{LLM-Flax}{Review DOI patch failed}{}}
\makeatother

\newcommand{\fref}[1]{Fig.~\ref{#1}}
\newcommand{\sref}[1]{Section~\ref{#1}}
\newcommand{\tref}[1]{Table~\ref{#1}}
\newcommand{\myparagraph}[1]{\noindent\textbf{#1}~}

\newcommand{\method}{LLM-Flax\xspace}

\newcommand{\fullmethod}{Full LLM-Flax\xspace}

\history{}
\doi{}

\title{LLM-Flax: LLM-Generated Rules and Object Guidance for Neuro-Symbolic Task Planning}

\author{%
\uppercase{Seongmin Kim}\authorrefmark{1} and
\uppercase{Daegyu Lee}\authorrefmark{2}%
}

\address[1]{Department of Physical AI Engineering, Jeonbuk National University (JBNU), Jeonju, Republic of Korea}
\address[2]{Department of Advanced Defense Technology and Industry, Jeonbuk National University (JBNU), Jeonju, Republic of Korea}

\corresp{Corresponding author: Daegyu Lee (e-mail: lee.dk@jbnu.ac.kr).}
\begin{document}

\begin{abstract}
Adapting a neuro-symbolic task planner to a new domain can require hand-authored symbolic rules and solved training instances for a learned object scorer.
This paper presents \method, which applies large language models (LLMs) to offline rule generation, failure recovery, and object scoring within the Flax planning pipeline.
Generated rules undergo validation and corrective feedback before reuse, while runtime guidance operates within a shared planning budget.
Experiments on MazeNamo and two additional domains show that the evaluated rule configuration improves planning success over manually authored rules while retaining the trained object scorer.
Across eight MazeNamo settings, the mean success rate increases by approximately three percentage points.
Controlled comparisons distinguish the effects of rule content from the choice of generator.
Runtime recovery and object scoring do not establish an overall performance advantage under the evaluated planning budgets.
These findings demonstrate the value of reusable LLM-generated rules and identify information quality and inference efficiency as central challenges for runtime guidance.

\end{abstract}


\titlepgskip=-15pt

\maketitle

\section{Introduction}\label{sec:introduction}
\IEEEPARstart{T}{ask} planning in complex, object-rich environments requires long-horizon reasoning over large state spaces, making classical symbolic planners computationally expensive~\cite{helmert2006fast}.
Neuro-symbolic methods reduce this search burden by using learned relevance estimates to select objects before invoking a symbolic solver~\cite{silver2021planning}.

Flax~\cite{du2026fast} follows this design by combining a graph neural network (GNN) object scorer with two forms of domain-specific symbolic knowledge: \emph{relaxation rules}, which remove selected objects to construct a simpler problem, and \emph{complementary rules}, which restore objects connected by spatial or structural relations.
The original study reports that these components reduce planning time on object-rich maze tasks.
Adapting Flax to another domain nevertheless introduces two distinct costs.
A domain expert must specify both rule types, and the GNN scorer requires a domain-specific set of solved planning instances.
The first cost concerns symbolic knowledge engineering; the second concerns data generation and model training.

Large language models (LLMs) support few-shot task specification and multi-step reasoning~\cite{brown2020language,wei2022chain}, motivating their use for constructing reusable domain rules.
We examine three uses of an LLM within Flax, separating offline rule generation from two forms of runtime guidance.

The resulting framework, \method, contains three stages:
\begin{itemize}[noitemsep,topsep=2pt]
    \item \textbf{Stage 1 --- LLM Rule Generation:} The LLM maps a Planning Domain Definition Language (PDDL) domain definition to relaxation and complementary rules. Schema validation and corrective prompting reject malformed outputs before planning.
    \item \textbf{Stage 2 --- LLM Failure Recovery:} After a timeout, the LLM selects excluded objects for one bounded replanning attempt. A feasibility check reserves time for the subsequent relaxation step.
    \item \textbf{Stage 3 --- LLM Zero-Shot Object Scoring:} The LLM assigns object-relevance scores from the current problem state and goal, replacing the domain-trained GNN for this zero-shot configuration.
\end{itemize}
Stage~1 addresses rule authoring while retaining the trained GNN. The original \fullmethod configuration combines the three interfaces without GNN labels; the new learned hybrid requires labels and is evaluated separately. All configurations require a PDDL domain and problem.
Our experiments evaluate the stages separately because they have different costs and levels of empirical support. Our contributions are an executable typed rule-generation interface, a multi-seed evaluation of rule-content effects against manual and PDDL-derived controls, and an end-to-end study of runtime recovery and scoring. These comparisons measure configuration performance, generation success and online utility as distinct outcomes.\footnote{Portions drafted or revised with OpenAI Codex~\cite{openai2026codex}; see Acknowledgments.}

\section{Related Work}
\label{sec:related}

\subsection{Neuro-Symbolic Task Planning}
Classical symbolic planners such as Fast Downward~\cite{helmert2006fast} and FF~\cite{hoffmann2001ff} search over formal state and action models, but grounding and search can become expensive as problem size grows.
STRIPS~\cite{fikes1971strips} established the foundational formalism, while PDDLStream~\cite{garrett2020pddlstream} extended it to continuous domains via blackbox samplers.
Neuro-symbolic approaches address scalability by coupling neural components with symbolic solvers.
PLOI~\cite{silver2021planning} introduced GNN-based object filtering, reducing the grounded state space and iteratively expanding the retained object set until a plan validates on the original problem.
Subsequent work explored neuro-symbolic skills~\cite{silver2022neurosymbolic}, relational transition models~\cite{chitnis2022learning}, predicate invention~\cite{silver2023predicate}, and efficient abstract planning models~\cite{kumar2023learning}.
LogiCity~\cite{li2024logicity} further advances neuro-symbolic artificial intelligence through large-scale abstract urban simulation.

Our work directly extends Flax~\cite{du2026fast}, which combines a GNN object scorer with hand-crafted relaxation and complementary rules for object-rich PDDL instances.
\method retains its symbolic planning loop and studies the rule-authoring and training-data costs as separate problems.

\subsection{LLMs for Planning and Formal Reasoning}
Prior work has used LLMs as planners, translators, and reasoning components.
SayCan~\cite{ahn2022can} grounds language-model outputs in robot affordances, while LLM+P~\cite{liu2023llm+} translates natural-language tasks into PDDL problems for an external solver.
Guan et al.~\cite{guan2023leveraging} use LLMs to construct or repair PDDL representations, whereas Tree-of-Thoughts~\cite{yao2023tree} and chain-of-thought prompting~\cite{wei2022chain} use language models to organize the reasoning process itself.

Stage~1 of \method instead produces a compact rule configuration once per domain and leaves plan generation to Fast Downward within the Flax loop.
Stages~2 and~3 query the LLM at runtime, but their outputs are restricted to object or cell identifiers, or scalar relevance scores, rather than actions or plans.
The Stage~1 validator limits structural errors before a rule file is reused across task instances; semantic errors remain possible and are evaluated separately in \sref{sec:quality}.

\subsection{Automated Knowledge Engineering}
Automated PDDL domain acquisition~\cite{cresswell2013acquiring} is an established problem in AI planning.
In contrast to learning a complete domain model from execution traces, Stage~1 maps an existing PDDL domain to the narrower JavaScript Object Notation (JSON) rule schema required by Flax~\cite{du2026fast}. This retains the existing domain model and solver, enabling domain-level rule reuse. Its limitation is that structural validity does not ensure an active or useful transformation, motivating separate acceptance, activity and planning-utility measurements.\footnote{Portions drafted or revised with OpenAI Codex~\cite{openai2026codex}; see Acknowledgments.}

\section{Methodology}\label{sec:method}
We first define the planning problem and the Flax execution loop, then specify typed rule generation and runtime object guidance.

\subsection{PDDL Planning}\label{sec:background}
A PDDL planning problem $\Pi = \langle \mathcal{D}, \tau \rangle$ consists of a domain $\mathcal{D} = \langle \mathcal{P}, \mathcal{A} \rangle$ defining lifted predicates and actions, and a task instance $\tau = \langle \mathcal{O}, I, G \rangle$ specifying objects, initial state, and goal.
A plan is a sequence of grounded actions mapping $I$ to a state satisfying $G$~\cite{edelkamp2004pddl2}.

\subsection{Neuro-Symbolic Planning with Rules}
\label{sec:flax_bg}
The baseline planner we extend, Flax~\cite{du2026fast}, operates in three steps during inference:

\myparagraph{Step 1 (GNN Pruning)}
A GNN assigns importance scores $s(o) \in [0,1]$ to each object. Starting from threshold $q = 0.81$ with decay $\gamma = 0.9$, the planner iteratively lowers $q$ and attempts to plan on the pruned object set $\mathcal{O}_1 = \mathcal O_G\cup\{o \mid s(o)>q\}$, where $\mathcal O_G$ contains goal objects, until success or timeout.

\myparagraph{Step 2 (Relaxation)}
When Step 1 yields no validated plan within its phase, \emph{relaxation rules} remove certain objects from the full problem to create a simplified ``rough'' version. The planner attempts to obtain a rough plan $\mu_r$; when one is found, its objects are merged back: $\mathcal{O}_2 = \mathcal{O}_1 \cup \{o \mid o \in \mu_r\}$. If rough planning fails without a timeout, Step~3 proceeds with the current retained set; a rough-planning timeout terminates the run.

\myparagraph{Step 3 (Complementary Expansion)}
\emph{Complementary rules} restore object pairs connected by structural constraints, yielding $\mathcal{O}_3 = \textsc{Comp}(\mathcal{O}_2)$, on which the final plan is computed.

Both rule types are stored as structured JSON configurations (see \sref{sec:method}) and loaded at runtime.
In the original system, these files are authored manually per domain.

\begin{figure*}[t]
  \centering
  \includegraphics[width=\textwidth]{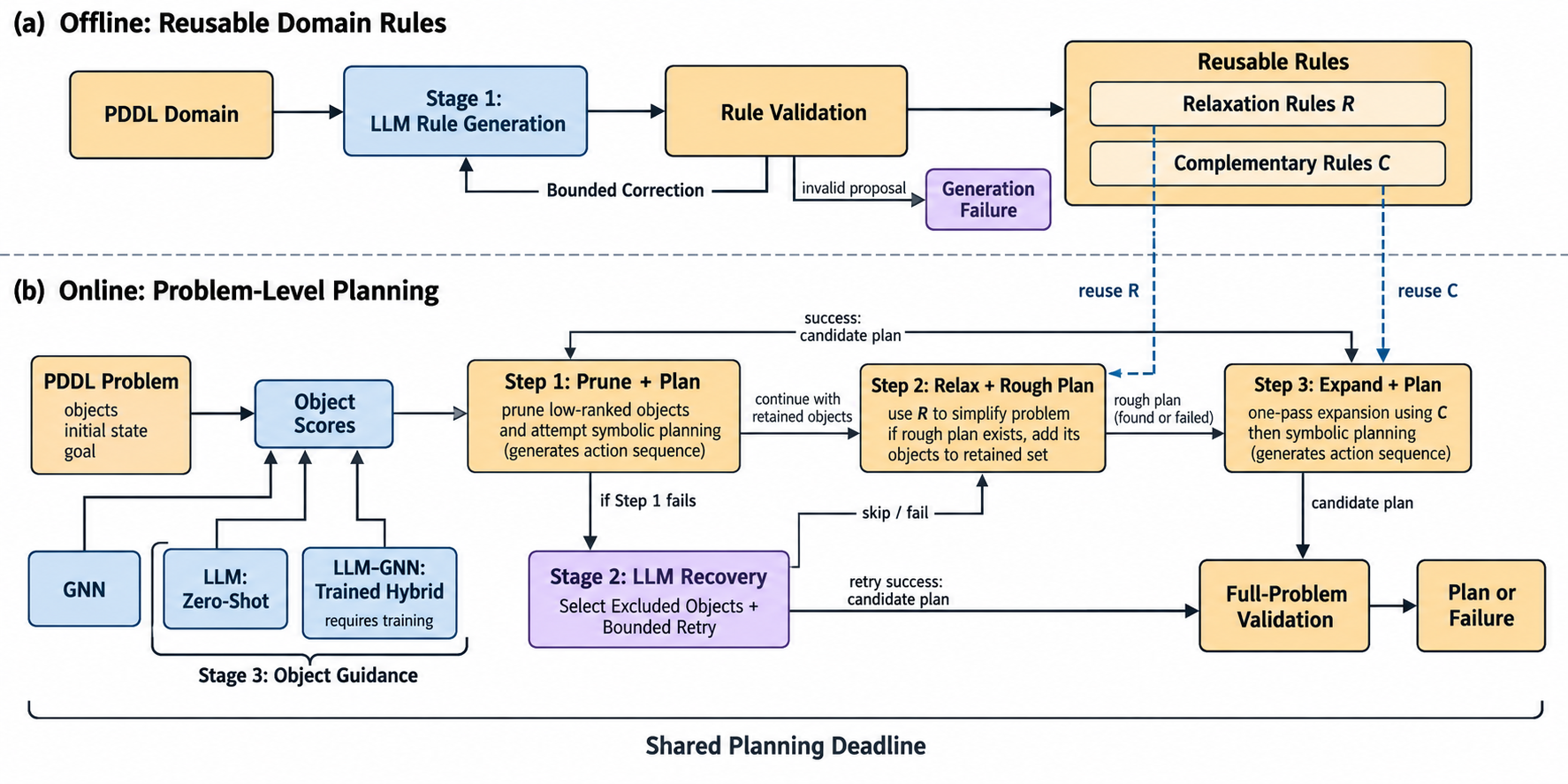}
  \caption{Architecture of \method. \emph{Top}: Stage~1 generates relaxation and complementary rules once per domain and validates the resulting JSON. \emph{Bottom}: the Flax planning loop performs threshold-decay pruning, rule-based relaxation, and complementary expansion. Stage~2 optionally inserts a feasibility-gated recovery call after a pruning timeout; Stage~3 provides zero-shot LLM or separately trained hybrid scoring as alternatives to GNN-only scores. Stage~1 proposes reusable rules; Stage~3 provides an alternative scorer. The learned hybrid requires labels. $\mathcal D=(\mathcal P,\mathcal A)$ is the domain, $\mathcal O,I,G$ are objects, initial state and goal, $s(o)$ is an object score, $U$ the retained set, $\mu_r$ a rough plan and $\pi$ a full-problem validated plan. FD is Fast Downward; VAL is the plan validator~\cite{howey2004val}. This schematic was generated with OpenAI image-generation tools~\cite{openai2026image}; see Acknowledgments.}
  \label{fig:architecture}
\end{figure*}

\subsection{Rule Format}
\label{sec:format}

\myparagraph{Relaxation rules}
Each rule specifies: (i) \texttt{pre\_compute}---a binary predicate used to look up a related object (e.g., \texttt{oAt} maps an obstacle to its position); (ii) \texttt{precond}---predicate(s) identifying removable objects; (iii) \texttt{delete\_objects}---binding slot of the object to remove; (iv) \texttt{delete\_effects}---literals to retract; (v) \texttt{add\_effects}---literals to assert after removal.
Indices are nonnegative shared variable-binding slots. A unary effect may refer to slot 1; these are not positions in every predicate. Complementary indices below instead refer to predicate arguments.

The manually crafted relaxation rule for MazeNamo reads:
\begin{minipage}{\columnwidth}
\begin{small}
\begin{verbatim}
{"rule0": {"pre_compute": {"oat":[0,1]},
  "precond": {"islight":[0]},
  "delete_objects": [0],
  "delete_effects": {"islight":[0],
    "ismoveable":[0], "oat":[0,1]},
  "add_effects": {"posempty":[1]}}}
\end{verbatim}
\end{small}
\end{minipage}

Semantically: \emph{for any light obstacle \texttt{o} at position \texttt{p}, remove \texttt{o} from the problem and mark \texttt{p} as empty.}
Let $S$ denote the current set of ground literals. For each consistent binding satisfying $\operatorname{isLight}(o)\land\operatorname{oAt}(o,p)$, define
\begin{equation}\begin{aligned}
 \mathcal O'&=\mathcal O\setminus\{o\},\\
 S'&=\{\ell\in(S\setminus D_{o,p})\cup\{\operatorname{posEmpty}(p)\}:\\
   &\hspace{4em}\operatorname{obj}(\ell)\subseteq\mathcal O'\},
\end{aligned}\label{eq:manual-rule}\end{equation}
where $D_{o,p}$ contains the three declared retractions above and $\operatorname{obj}(\ell)$ is the set of objects in literal $\ell$.
The transformation is skipped atomically if $o$ is a goal object; optional exclusion predicates further protect matched objects.

\myparagraph{Complementary rules}
Each entry maps a binary predicate name to paired \texttt{cond}/\allowbreak\texttt{cmpl} index lists.
For each paired index set $(J,K)$, all objects indexed by $J$ must be in the input set $U$ before objects indexed by $K$ are added:
\begin{equation}
 C(U)=U\cup\bigcup_{p(\mathbf o)\in S}\ \bigcup_{(J,K)\in C_p:\{o_j:j\in J\}\subseteq U}\{o_k:k\in K\}.
\label{eq:complementary}\end{equation}
Here $C_p$ is the set of paired index sets for predicate $p$. Conditions are tested against the same input $U$ in one pass, not a transitive closure.
The manually crafted complementary rule is simply \texttt{\{"oat": \{"cond":[[0],[1]], "cmpl":[[1],[0]]\}\}}.

\subsection{LLM Rule Generation}
\label{sec:generation}

We use Gemma3-12B~\cite{gemma3_2025}, hosted locally through Ollama's application programming interface (API), for the fixed Stage~1 rule configuration.
The original generation uses temperature~0 and a 16,384-token context window.
The model is queried twice per domain, once for each rule type.
Fresh-generation reliability is evaluated separately with the matched-prompt local/hosted protocol in \sref{sec:model_comparison}.

\myparagraph{Prompt design}
Each historical prompt defines the rule semantics and JSON schema, states that argument indices must be integers, and provides one valid example and one counterexample with string-valued indices.
The complete PDDL domain definition is included so that predicate names and arities are available to the model.
The new component-isolated prompts instead use typed named variables, contain no manual-rule examples or solved plans, and rank a fixed PDDL-derived pool for complementary selection.

\myparagraph{Validation and self-correction}
LLM outputs are validated by \textsc{ValidateRules}, which checks: (i) all required keys are present; (ii) all index fields contain lists of integers; (iii) all predicate names exist in the domain (parsed from the PDDL \texttt{:predicates} block); and (iv) arities, binding slots and types are consistent.
The revised compiler also rejects unsupported duplicate-atom mappings and unbound slots. Complementary indices are argument positions; relaxation indices are shared binding slots.
If validation fails, the error messages are fed back to the LLM as a correction turn, with up to $K=3$ total attempts.

\myparagraph{Post-processing}
Rules with identical compiled content are deduplicated; unknown or malformed constructs are rejected rather than silently filtered after outcome inspection.
The accepted rules are then saved as JSON files and loaded by the planner.
Raw outputs, errors, call latency and generation cost are retained, including failed sessions and accepted but inactive proposals.
Activity is measured separately from structural acceptance by checking whether a proposal changes an inspected development state.

Algorithm~\ref{alg:gen} summarizes Stage~1.

\subsection{Stage 2: LLM-Guided Failure Recovery}
\label{sec:stage2}

Step~1 uses threshold decay to expand the selected object set after an unsuccessful planning attempt.
This procedure does not distinguish among the excluded objects.

We introduce \textsc{LLMRecoveryGuidance}, inserted between a Step~1 timeout and the Step~2 relaxation fallback.
Given the PDDL state, goal, included objects $\mathcal{O}_{cur}$, and excluded candidates $\mathcal{O}_{exc}$, the LLM selects excluded objects that may be needed for a plan.
The LLM returns a JSON list of object names; the planner adds them and retries.
If the retry succeeds the plan is returned; otherwise the pipeline continues to Step~2 with its current object set.

\myparagraph{Budget policy}
LLM API calls impose a latency cost that must be accounted for in per-problem timeout budgets.
We adopt a three-part policy (see \sref{app:budget} for its budget analysis):
\begin{itemize}[noitemsep,topsep=2pt]
  \item \emph{Feasibility check}: reserve 5\,s for Step~2 and at least 1\,s for retry, allowing an LLM call of at most 5\,s within the remaining Step~2 slack.
  \item \emph{Recovery cap}: limit the recovery replan to $15\%$ of total timeout.
  \item \emph{Step~2 reservation}: when recovery is feasible, retain at least $5$\,s for the subsequent relaxation step.
\end{itemize}
All online operations share a monotonic deadline; pruning ends at $T/6$ and rough planning at $T/2$.
API calls use actual timeouts with no automatic retry. No request or replanning attempt starts after its deadline, and every returned plan must pass full-problem VAL within the total budget.
The legacy policy retains the 11\,s gate; the revised policy permits shorter calls.
Goal objects are always retained, object ordering is fixed and threshold decay explicitly includes equal/zero-score objects before termination.

\subsection{Stage 3: LLM Zero-Shot Object Scoring}
\label{sec:stage3}

The MazeNamo cache contains 195 successful planning labels from 200 attempts.
Stage~3 removes this training requirement by replacing the GNN with \textsc{LLMObjectGuidance}, which implements the same \texttt{score\_object(obj, state)} interface.

On the first call for a given state, the LLM receives the goal, the list of all objects with types, and up to 80 state facts, and returns a JSON dictionary mapping each object name to a score in $[0, 1]$.
Direct scores are cached, so this direct variant makes one LLM call per planning problem; batching can make additional bounded calls.
The \texttt{train()} method performs no data collection or parameter updates.
The revised batched variant scores 32 sorted objects at a time with at most one omission-correction call per batch; a separate variant supplies goal-related and batch-object facts.
Fallback values 0.1/0.3/0.5/0.7/0.9 are compared by replaying identical raw outputs, separately from end-to-end inference measurements.

If the call fails, or if an object is absent from the returned dictionary, its score defaults to 0.5.

\myparagraph{Full LLM-Flax}
\fullmethod combines the three stages: Stage~1 generates rules from the domain definition, and Stages~2 and~3 provide runtime recovery and object scores.
This zero-shot configuration uses fixed LLM-generated rule files and no GNN labels; the hybrid below is a separately trained scorer.
\myparagraph{Learned hybrid extension}
A distinct hybrid augments the existing GNN input with an LLM score/hint channel and a valid-response channel:
$s_\phi(o)=\operatorname{sigmoid}(f_\phi(x(o),z(o),m(o)))$.
Only GNN parameters $\phi$ are trained on reference-plan object membership; the LLM is frozen. These labels do not certify object necessity.
Small-label experiments use nested budgets of 20/50 successes including 10\% validation, keeping architecture, loss, optimizer and 301-epoch final-checkpoint selection fixed.
Training features are cached; actual LLM calls are included at evaluation. Finite/nonzero gradients, parameter updates, tiny-set overfit and learning curves are checked before interpreting performance.
The later full-data experiment uses compact cell hints and a global valid-response channel rather than dense object scores, with masked, shuffled and separately trained constant-feature controls.
The hybrid requires labels and online inference and is not zero-shot.\footnote{Portions drafted or revised with OpenAI Codex~\cite{openai2026codex}; see Acknowledgments.}

\begin{algorithm}[t]
\caption{\method Rule Generation (Stage 1)}\label{alg:gen}
\SetKwInOut{Input}{Input}\SetKwInOut{Output}{Output}
\Input{PDDL domain $\mathcal D$; component $q\in\{R,C\}$; model/settings $\mathcal M$; schema; $K=3$; resource budgets}
\Output{Accepted component $r_q$ and trace, or unavailable reason}
\textbf{Precondition:} predicate/type signatures and schema are fixed\;
$H\leftarrow[\textsc{BuildPrompt}(\mathcal D,q)]$; $r_q\leftarrow\bot$\;
\For{$k\leftarrow1$ \KwTo $K$}{
 \If{request/resource budget exhausted}{\Return unavailable (budget)\;}
 $x\leftarrow\mathcal M(H)$ with explicit timeout and no automatic retry\;
 \If{fatal transport or accounting failure}{record reason; \Return unavailable (request failure)\;}
 save raw output, identity, latency and usage\;
 $(r,e)\leftarrow\textsc{ParseCompileValidate}(x,\mathcal D,q)$\;
 \If{$e=\emptyset$}{record activity separately; \Return $r$ and trace\;}
 $H\leftarrow H\mathbin{\|}(x,e)$\;
}
\Return unavailable (recorded validation failures)\;
\end{algorithm}

\section{Experiments}
\label{sec:experiments}

\subsection{Setup}\label{sec:setup}

\myparagraph{Domain}
We evaluate on the MazeNamo benchmark~\cite{du2026fast}: a grid-based maze navigation task where a robot must reach a goal while manipulating heavy boxes (push-only) and light boxes (push or pick-and-place).
The eight original settings are retained: $10{\times}10$ easy/medium/hard, $12{\times}12$ medium/hard/expert and $15{\times}15$ medium/hard.
The revised fixed-rule and relation-ablation comparisons use IDs 0--49 in every relevant folder, with ten GNN seeds; budgets remain 10/30/40\,s by grid size.
The cached MazeNamo training pool has 195 successful labels from 200 attempts, split 176/19 for training/validation.
The scorer uses weighted binary cross entropy (positive weight 10), Adam at $10^{-3}$ and the final checkpoint after 301 epochs.
SokomindPlus and DifficultLogistics use domain-trained GNNs and five seeds (\sref{sec:crossdomain_results}).

\myparagraph{Configurations}
Stage~1 is evaluated on all eight MazeNamo benchmarks. Stages~2 and~3 are evaluated on four selected difficult benchmarks: 12 Hard/Expert and 15 Medium/Hard.
Stage~1 compares two rule configurations, holding all other components fixed:
\begin{itemize}[noitemsep,topsep=2pt]
    \item \textbf{Manual}: Hand-crafted rules (1 relaxation, 1 complementary). Timeouts: 10\,s / 30\,s / 40\,s for $10{\times}10$ / $12{\times}12$ / $15{\times}15$.
    \item \textbf{\method}: LLM-generated rules (1 relaxation after dedup, 7 complementary). Same timeouts.
\end{itemize}
The fixed rules are the existing Gemma3-12B files, evaluated without regeneration; fresh-generation reliability is measured separately.
Stage~2 adds \textbf{\method + Recovery}, using \textsc{LLMRecoveryGuidance} with the budget policy in \sref{sec:stage2}, alongside legacy and equal-count random controls.
Stage~3 compares \textsc{LLMObjectGuidance} (zero-shot), bounded batches and separately label-trained hybrids while keeping the rules fixed. Recovery is disabled to isolate the scorer; these comparisons do not evaluate the three-stage \fullmethod combination.
\sref{sec:crossdomain_results} evaluates Stage~1 with domain-trained GNNs. Hosted rule comparisons hold one component fixed; a fully generated GPT $(R,C)$ combination is not measured.
Table~\ref{tab:configuration-overview} summarizes the rule, scorer, label and online-call requirements.

\begin{table}[tbp]\centering\scriptsize
\caption{Principal configurations. Labels count successful training instances, including validation; $N=195$ for MazeNamo, 179 for SokomindPlus and 198 for DifficultLogistics. The hybrid row denotes the full-data study; 20/50-label variants are in the supplement. Online calls exclude offline rule generation.}\label{tab:configuration-overview}
\setlength{\tabcolsep}{2.5pt}
\begin{tabular}{@{}p{18mm}p{10mm}p{12mm}p{8mm}p{10mm}p{11mm}@{}}\toprule
Configuration & Rule source & Scorer & Labels & Recovery & Online LLM\\\midrule
Pure FD & None & None & 0 & Off & No\\
PLOI & None & GNN & $N$ & Off & No\\
Manual Flax & Manual & GNN & $N$ & Off & No\\
LLM-Flax (S1) & Gemma & GNN & $N$ & Off & No\\
S1 + recovery & Gemma & GNN & $N$ & Gated & Gated\\
S3 zero-shot & Gemma & LLM & 0 & Off & Yes\\
S3 hybrid & Gemma & GNN + LLM & $N$ & Off & Yes\\
\bottomrule\end{tabular}\end{table}

\myparagraph{Cohorts and exposure}
IDs 0--49 and the previously used development sets are now exposed revision evidence.
The hosted pilot uses ten IDs 50--59 in each of 12 Hard/Expert and 15 Medium/Hard, with fixed GNN seed 0 and five generation sessions per model/component.
Earlier typed-candidate comparisons use 20 development maps and 30 separately registered maps per setting; their completed cohorts are retained as historical evidence.
The primary full-data runtime cohort instead contains 300 newly sampled, unfiltered maps at three sizes, with ten GNN seeds. Its state/file overlap audit checks training and recorded previous evaluations; untraced historical exposure cannot be excluded.
These distributions are reported separately: original difficulty buckets were conditioned on reference-planner runtime, whereas the new maps are unfiltered.

\myparagraph{Metrics and computation}
Success rate (SR) is the fraction of problems solved by full-problem VAL-validated plans within the total deadline. We report seed mean and sample standard deviation (SD), 10,000-sample crossed paired seed/problem bootstrap 95\% confidence intervals (CIs), success-only time $t_s$, failure-capped all-problem time $t_c$, and paired common-success time.
For failure, $t_c$ assigns the full timeout; a fast failure is not a speedup.
The hosted pilot instead crosses generation sessions and maps within fixed settings; duplicate compiled artifacts are shared observations, not independent replications. Its intervals are exploratory and unadjusted.
Unless stated otherwise, planning time covers the complete planner call and therefore includes runtime LLM latency in Stages~2 and~3. Every request, solver call and final validation is charged to the planning deadline. Model loading/environment setup is outside the measured planner call; offline generation/training costs are reported separately.
All evaluation configurations use Fast Downward (FD) with \texttt{lama-first}; the local machine has an Intel Core Ultra 9 285K and RTX 5080 (16,303 MiB, driver 580.95.05).
The matched local pilot uses Ollama 0.21.0 and Q4\_K\_M weights. The principal local model tags are \texttt{gemma3:12b} and \texttt{qwen2.5:14b}; the hosted tag is \texttt{gpt-6-astra}. The Supplement explains how to locate the saved prompts, settings, rules and checkpoint identifiers. Package versions, prompts, checkpoint identifiers, rule/data hashes and recorded machine load are included with the supplementary reproducibility records. Model weights and the full problem corpus are separate from this submission archive.

\subsection{Results}
\label{sec:results}
\begin{table*}[tbp]\centering\small
\caption{Reference methods evaluated in our pipeline on 50 maps per setting. SR is seed mean $\pm$ SD; $t_c$ is failure-capped mean seconds. GNN methods use ten seeds for MazeNamo and five for the additional domains; Pure FD has one execution per map. Success-only times and plan-length records are supplied in the supplement.}\label{tab:main}\label{tab:fixed0}\label{tab:fixed16}\label{tab:baselines}\label{tab:crossdomain}\label{tab:fixed32}
\setlength{\tabcolsep}{3pt}
\begin{tabular}{lrrrrrrrr}\toprule
 & \multicolumn{2}{c}{Pure FD} & \multicolumn{2}{c}{PLOI~\cite{silver2021planning}} & \multicolumn{2}{c}{Manual Flax~\cite{du2026fast}} & \multicolumn{2}{c}{LLM-Flax (S1)}\\\cmidrule(lr){2-3}\cmidrule(lr){4-5}\cmidrule(lr){6-7}\cmidrule(lr){8-9}Task & SR (\%) & $t_c$ & SR (\%) & $t_c$ & SR (\%) & $t_c$ & SR (\%) & $t_c$\\\midrule
10 Easy & 100.0 & 0.56 & 95.8 $\pm$ 1.5 & 1.00 & 99.2 $\pm$ 1.9 & 0.71 & 99.6 $\pm$ 0.8 & 0.68\\
10 Medium & 100.0 & 1.21 & 91.6 $\pm$ 3.1 & 1.46 & 96.8 $\pm$ 1.4 & 0.96 & 98.4 $\pm$ 0.8 & 0.85\\
10 Hard & 100.0 & 2.74 & 84.6 $\pm$ 2.7 & 2.38 & 93.8 $\pm$ 2.0 & 1.71 & 96.2 $\pm$ 1.1 & 1.58\\
12 Medium & 100.0 & 6.24 & 83.2 $\pm$ 2.7 & 6.40 & 95.8 $\pm$ 1.8 & 3.25 & 99.2 $\pm$ 1.0 & 2.57\\
12 Hard & 100.0 & 17.37 & 60.4 $\pm$ 4.5 & 13.08 & 87.2 $\pm$ 3.0 & 7.20 & 95.0 $\pm$ 2.4 & 6.14\\
12 Expert & 42.0 & 28.63 & 48.2 $\pm$ 4.8 & 16.87 & 69.0 $\pm$ 3.3 & 13.02 & 74.6 $\pm$ 2.3 & 12.78\\
15 Medium & 94.0 & 28.08 & 75.6 $\pm$ 4.3 & 12.39 & 91.8 $\pm$ 2.7 & 8.92 & 93.8 $\pm$ 1.5 & 9.68\\
15 Hard & 38.0 & 37.03 & 71.6 $\pm$ 4.3 & 13.64 & 94.0 $\pm$ 2.5 & 8.99 & 94.6 $\pm$ 3.0 & 9.88\\
SokomindPlus & 100.0 & 13.08 & 43.2 $\pm$ 6.4 & 19.82 & 64.4 $\pm$ 6.1 & 15.40 & 86.8 $\pm$ 4.1 & 12.77\\
DifficultLogistics & 94.0 & 42.30 & 97.2 $\pm$ 1.8 & 22.29 & 82.8 $\pm$ 9.0 & 31.96 & 92.4 $\pm$ 4.1 & 28.22\\
\bottomrule\end{tabular}\end{table*}

\begin{figure}[tbp]
    \centering
    \includegraphics[width=\columnwidth]{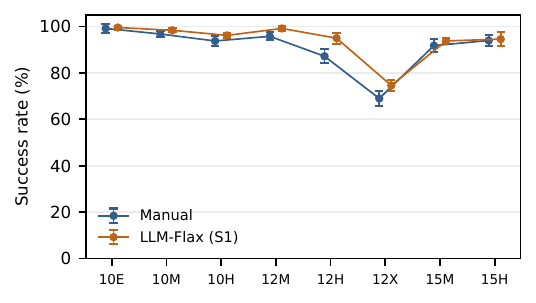}
    \caption{Updated fixed-configuration SR across the original eight MazeNamo settings (E/M/H/X: easy/medium/hard/expert). Error bars are sample SD across ten GNN seeds on 50 maps per setting. The same scorer, deadlines and corrected planner are used; fixed LLM-generated rules are distinct from fresh generation.}
    \label{fig:sr_scale}
\end{figure}

\begin{figure*}[!t]
  \centering
  \includegraphics[width=0.85\textwidth]{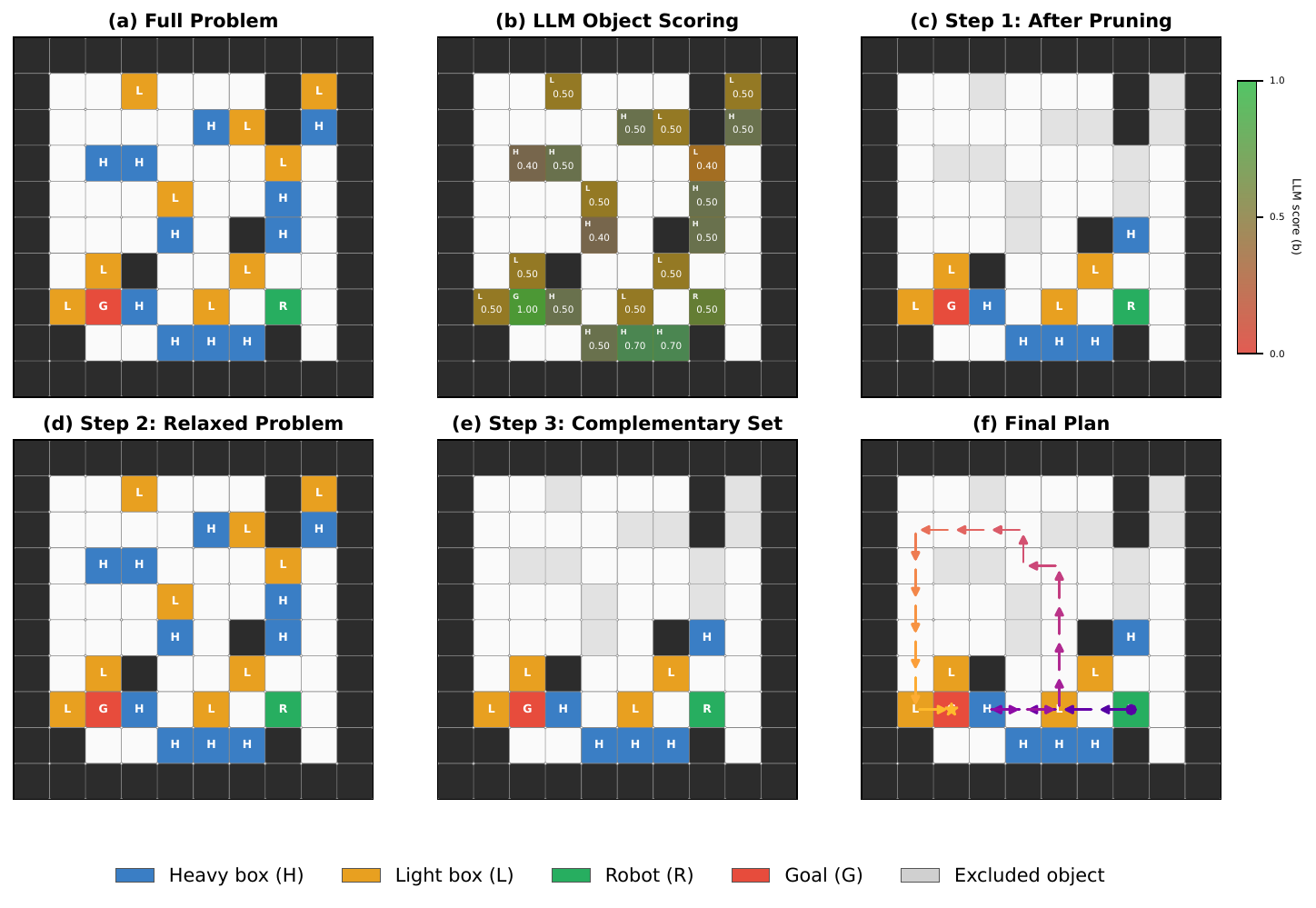}
  \caption{Illustrative trace using LLM-generated rules and precomputed zero-shot LLM scores on a $10{\times}10$ hard MazeNamo problem (problem~16, 163 objects).
  (a)~Full problem with all objects.
  (b)~Zero-shot relevance scores used for this trace.
  (c)~After Step~1 threshold-based pruning: 21 objects excluded (grey), 142 retained at threshold 0.478.
  (d)~Step~2 relaxed sub-problem: light boxes removed via relaxation rules, yielding a rough plan.
  (e)~Step~3 complementary expansion: object pairs with structural constraints restored.
  (f)~Final plan of length 39. This is an archived qualitative trace from the submitted paper, preserved for illustration. Its 0.8\,s excludes score generation and is not a current end-to-end Stage~3 result or evidence for the revised numerical tables.}
  \label{fig:pipeline}
\end{figure*}

\tref{tab:main} reports the Stage~1 comparison, and \fref{fig:sr_scale} summarizes its SR values. \fref{fig:pipeline} provides a separate qualitative trace of the zero-shot scoring interface.

The observed SR with LLM-generated rules is higher than that with manual rules on each benchmark.
The unweighted mean across the eight tasks is 93.93\% for LLM-generated rules and 90.95\% for manual rules.
The largest mean difference is on 12 Hard: 87.2\% to 95.0\% (+7.8 percentage points; paired 95\% interval [+2.4,+15.0]).
On 15 Hard, SR changes only from 94.0\% to 94.6\%, while failure-capped mean time increases from 8.99 to 9.88\,s.
On common solved 12 Hard problem/seed pairs, LLM-Flax planning is 0.42\,s slower, despite a lower all-problem capped mean; success and time therefore require joint interpretation.
These fixed-rule differences identify rule-content effects, not a generator-specific advantage. All per-setting paired intervals and common-success times are supplied in the supplement.

\begin{figure}[tbp]\centering
\includegraphics[width=\columnwidth]{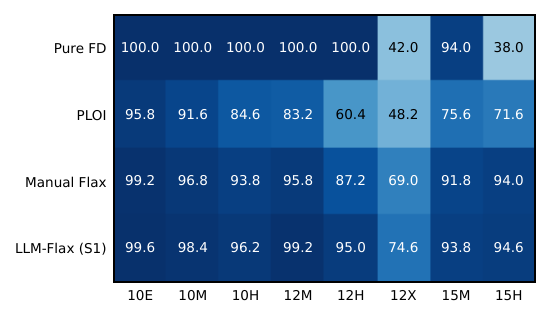}
\caption{Matched reference-method comparison across the original eight MazeNamo settings. Values are SR (\%); FD has one execution per map and other rows are means over ten GNN seeds. Complete variability and capped times are in Table~\ref{tab:main}. Earlier unmatched LLM-generator rows are not pooled into this comparison.}
\label{fig:sr_heatmap}\end{figure}

\subsection{Rule Quality Analysis}
\label{sec:quality}

\myparagraph{Relaxation rules}
Gemma3-12B generates the same MazeNamo relaxation rule as the manual configuration in the fixed files.
The \texttt{pre\_compute}, \texttt{precond}, \texttt{delete\_objects}, \texttt{delete\_effects}, and \texttt{add\_effects} fields are identical, so differences in the main comparison arise from the complementary rule sets or run-to-run planning behavior.
The new typed-generation study separates structural acceptance, actual state transformations and planning utility (\sref{sec:model_comparison}).
The earlier per-model rule inspection is retained in the supplement as historical diagnostics rather than an outcome-matched ranking.

\myparagraph{Complementary rules}
Gemma3-12B generates seven complementary rules in the fixed configuration (\texttt{upon}, \texttt{oat}, \texttt{rat}, \texttt{upto}, \texttt{downto}, \texttt{leftto}, \texttt{rightto}) versus manual \texttt{oat}.
Because rule count and predicate coverage change together, the main comparison alone cannot attribute the SR difference to either factor.
We therefore add each of the other six predicates individually to \texttt{oat}, and remove each from the full set, on 12 Expert and 15 Hard with ten GNN seeds and 50 maps.
Relaxation, checkpoints and problems are fixed. All 28 endpoint/addition/removal cohorts are complete.
The oat-only/full endpoints are 69.2/75.0\% on 12 Expert and 93.8/94.8\% on 15 Hard; full seed SD and timing appear with all 28 ablation cohorts in the supplement. These endpoints were executed separately from the fixed-rule rows in Table~\ref{tab:main}; they are not reused observations. Matched problem/checkpoint hashes and parsed rule objects agree. The 0.2--0.4 percentage-point offsets are compatible with timing sensitivity under wall-clock deadlines; no single runtime cause was isolated. We use within-ablation contrasts rather than pooling the repeated endpoints.

The endpoint difference is +5.8 percentage points on 12 Expert and +1.0 on 15 Hard; individual additions/removals and their paired comparisons are in the supplement.
A separate read-only audit finds identical grounded complementary implications between the stored seven-predicate rules and a generic all-binary enumerator on 195 training plus 160 development states.
The extra generic \texttt{holding} relation has no facts in those 355 states, implying identical one-pass expansion for every input subset there.
This audit identifies the same one-pass expansion on the inspected states; arbitrary unseen and intermediate states remain outside the equivalence check.

\subsection{Stage 2: LLM-Guided Failure Recovery}
\label{sec:results_stage2}

The revised evaluation covers 12 Hard/Expert and 15 Medium/Hard, using 50 maps per setting and ten GNN seeds with a fixed stored rule fixture. Recovery uses Qwen2.5 14B~\cite{qwen2025qwen25} with temperature 0, seed 0 and a 16,384-token context.
\begin{table}[tbp]\centering\small
\caption{Recovery SR (\%, mean $\pm$ SD), 50 maps and ten GNN seeds per setting. Legacy-policy, timing and call/object diagnostics are in Supplement Sec. 2. H/X/M denote hard/expert/medium.}\label{tab:stage2}\label{tab:recovery0}
\setlength{\tabcolsep}{3pt}
\begin{tabular}{lrrr}\toprule
Task & None & Bounded & Random\\\midrule
12H & 95.0 $\pm$ 2.4 & 93.8 $\pm$ 2.7 & 94.6 $\pm$ 2.3\\
12X & 74.6 $\pm$ 2.3 & 71.8 $\pm$ 2.7 & 72.2 $\pm$ 2.6\\
15M & 93.8 $\pm$ 1.5 & 93.0 $\pm$ 1.7 & 93.0 $\pm$ 1.7\\
15H & 94.6 $\pm$ 3.0 & 92.2 $\pm$ 3.3 & 92.6 $\pm$ 3.5\\
\bottomrule\end{tabular}\end{table}

Bounded recovery issues 879 calls at 879 recorded opportunities across 2,000 problem/seed executions, producing 27 immediate retry successes.
These do not establish overall SR gain: opportunities already have a subsequent symbolic fallback, and the equal-count random control is competitive.
The legacy gate skips all 522 opportunities in the two 12-cell settings for insufficient Step 2 slack.
The supplement reports the complete legacy comparison, failure-capped times, proposal/adoption counts and final-plan occurrence with its successful-call denominator; appearing in a returned plan is not proof of object necessity.

\subsection{Validation Safeguards}
\label{sec:ablation_pipeline}\label{sec:historical}

The revised executor fixes the inverted condition for single-direction complementary entries and joins relaxation preconditions conjunctively. Goal-object protection and retained-object literal filtering are explicit.
A shared monotonic deadline, actual API timeout, disabled automatic retries and a descendant-process watchdog bound runtime. Every quantitative success must pass VAL on the full original problem.
All revised methods share these corrections; pre-correction submitted numbers are not mixed into new comparisons.

The earlier four-model/three-domain prompt interventions accepted 1/12, 3/12 and 5/12 domain/model pairs for prompt, schema and filtered-schema procedures; all MazeNamo pairs failed, and four of the last five accepted pairs had only no-op complementary directions.
The historical typed-candidate pool contains 76 structurally accepted proposals and 42 development-active proposals, from sequential candidates/corrections within one session per domain/model, not 76 independent generations.
Of 124 registered final cohorts, 90 are complete and 34 incomplete; only complete grids are reported. The combined pool-selected method scores 60.0\%/85.0\%/67.3\% on the complete 30-map 12 Expert/15 Hard/SokomindPlus cohorts, versus manual 76.3\%/95.3\%/66.7\%.
These results and all complete non-LLM controls are retained in the supplement; incomplete logistics comparisons are not filled with zero outcomes.

\subsection{LLM Model Comparison (Stage 1)}
\label{sec:model_comparison}\label{sec:hosted}

The original mixed-setting model ranking is replaced by a repeated-generation, component-isolated comparison.
Gemma3 12B and Qwen2.5 14B~\cite{qwen2025qwen25} provide five archived generation sessions each for complementary selection (C) and relaxation (R); GPT-6 Astra contributes five new sessions per component.
All resulting planning configurations and non-LLM controls are freshly evaluated on the same 40 exposed maps, GNN seed 0, compiler, FD settings and 30/40\,s budgets.
Exact compiled $(R,C)$ pairs are deduplicated: 25 unique configurations and 89 source aliases give 1,000 actual problem executions, with all 848 successes passing full-problem VAL.

\myparagraph{Inference settings}
Local generation uses Q4\_K\_M models, temperature 0.2, seeds 0--4, context 8,192, C/R output limits 512/1,536 and 90\,s per request.
The hosted identifier is \texttt{gpt-6-astra}, high reasoning effort and standard tier, with a 25,000-token cap including reasoning and 180\,s per request; API seed/temperature controls are not used.
Semantic prompts match the local pilot byte-for-byte. A documented provider-schema adapter represents disjoint predicate alternatives using \texttt{anyOf}; the same local validator enforces the semantics and ranking uniqueness.
Thus model, provider, quantization, decoding and resource budget differ together; this is not a causal parameter-count experiment.
The registered hosted cap is 30 requests/USD~20; the actual ten first-call successes consume about 267.0\,s of request time and an uncached-rate token-cost upper estimate of USD~0.994.
This is offline machine cost, not measured human labor.

All C sessions are accepted on the first call. The complete six-cell generation/first-call/distinct-output breakdown is retained in the supplement.

\myparagraph{Relaxation generation}
With all-binary C fixed, GPT accepts five of five R sessions on the first call and all five transform all 40 inspected initial states.
Gemma accepts none; Qwen accepts two after corrections, both inactive.
All five GPT sessions succeed on the first call and produce two distinct compiled R outputs. The exact 95\% acceptance interval is [47.8,100]\%, reflecting the five-session sample size.

\begin{table}[tbp]\centering\small
\caption{Relaxation generation and conditional planning utility on 40 exposed maps, GNN seed 0, with all-binary C fixed. Acc./Act. are accepted/active sessions; SR is percent and $t_c$ seconds. Failed generations have no fabricated planning score.}\label{tab:hosted-r}\label{tab:model_comparison}
\setlength{\tabcolsep}{3pt}
\begin{tabular}{lrrrr}\toprule
R source & Acc. & Act. & SR & $t_c$\\\midrule
Gemma3 12B & 0/5 & 0 & --- & ---\\
Qwen2.5 14B & 2/5 & 0 & 62.5 & 13.07\\
GPT-6 Astra & 5/5 & 5 & 90.0 & 7.68\\
Manual & --- & --- & 90.0 & 7.62\\
PDDL-action & --- & --- & 90.0 & 7.74\\
\bottomrule\end{tabular}\end{table}

Accepted GPT R proposals reach 90.0\% SR, matching manual and action-derived R, versus 62.5\% for the two accepted Qwen proposals.
The GPT common guard is
\begin{equation}\begin{aligned}
 &\operatorname{isMoveable}(o)\land\operatorname{onGround}(o)\land\operatorname{clear}(o)\\
 &\quad\land\operatorname{oAt}(o,p)\land\neg\operatorname{isHeavy}(o).
\end{aligned}\label{eq:gpt-guard}\end{equation}
The rule removes $o$ and asserts $\operatorname{posEmpty}(p)$.
A read-only posthoc audit finds that both variants produce exactly the manual reference's retained objects, relaxed literals and goals on these 40 initial states.
The audit establishes manual-reference behavior on the inspected states and explains the matched planning SR.

\myparagraph{Complementary selection}
Each C session ranks the same eight binary predicates; prefixes of 1/2/4 predicates give 2/4/8 directed implications with manual R fixed.
Controls use manual \texttt{oat}, all binary predicates, a ranking by distinct PDDL-operator occurrence (lexical tie), and five fixed random rankings.
The generic R control is selected by development activity, minimum atom count and a canonical-ID tie, without planning labels.

\begin{table}[tbp]\centering\small
\caption{Matched C-selection SR (\%) with manual R fixed; $k$ predicates give $2k$ directions. Means include all five sessions/rankings per source. Manual oat alone scores 90.0\%.}\label{tab:hosted-c}
\setlength{\tabcolsep}{3pt}
\begin{tabular}{rrrrrr}\toprule
$k$ & Gemma & Qwen & GPT & Action & Random\\\midrule
1 & 82.5 & 82.5 & 84.0 & 82.5 & 84.0\\
2 & 90.0 & 90.0 & 90.0 & 90.0 & 85.5\\
4 & 90.0 & 90.0 & 90.0 & 90.0 & 85.5\\
\bottomrule\end{tabular}\end{table}

GPT's C1 mean is 84.0\% versus action's 82.5\% (exploratory paired difference +1.5 percentage points, interval [0,+6]); four of five GPT rankings are weaker than its best ranking.
At C2/C4 the compiled GPT choices coincide with action controls and both reach 90.0\%; manual \texttt{oat} alone already reaches 90.0\%.
All five sessions are reported, without choosing the best output after observing SR.
The hosted configuration resolves the observed local R-generation validity/activity failures and achieves the manual/action reference SR. The matched-budget results identify working rule generation and selection as the measured contribution.

The supplementary hosted-model figure provides the complete budget and acceptance/activity visualization.

\subsection{Rule Transfer to Additional Domains}
\label{sec:crossdomain_results}

We evaluate Stage~1 on two additional PDDL domains in the Flax codebase. \textbf{SokomindPlus} is a Sokoban-style box-pushing domain that shares several relations with MazeNamo, whereas \textbf{DifficultLogistics} models trucks, airplanes, packages, locked locations, and stacking.
Both domains ship with manual rules and separate training folders. The revision collects 179 SokomindPlus and 198 DifficultLogistics successful labels from 200 attempts each, then trains domain-specific GNNs for five seeds.
Label collection takes 3,106.9\,s and 624.4\,s respectively and is separate from online planning.
The earlier random-scorer probes remain historical comparisons; the updated quantitative comparison in Table~\ref{tab:main} uses trained scorers and 50 maps per domain.
Hyphenated predicate parsing, unsolvable-rough-plan fallback and the DifficultLogistics free-ground predicate correction are shared by all revised rule configurations.

Table~\ref{tab:main} includes all four baseline methods for both additional domains.

\myparagraph{SokomindPlus: Stage~1 transfer}
LLM-Flax reaches 86.8\% versus manual 64.4\% (+22.4 percentage points; paired interval [+11.6,+33.2]). Pure FD reaches 100.0\%.
\myparagraph{DifficultLogistics}
LLM-Flax reaches 92.4\% versus manual 82.8\% (+9.6 points; interval [+0.4,+19.6]); PLOI reaches 97.2\% and Pure FD 94.0\%.
Thus a favorable rule configuration need not be the strongest planner, and these three domains do not establish general cross-domain generation utility.

\myparagraph{A flexibility case study: extending the rule schema itself}
\label{sec:exclude_precond}
The DifficultLogistics failure analysis identifies a limitation in the original rule schema. A broad \texttt{precond}, such as \texttt{obj}, can also match a gating object marked by a narrower predicate such as \texttt{key-package}; removing that object can break a global dependency.
We added an optional \texttt{exclude\_precond} field to represent ``remove X unless X is also Y.'' With a domain-independent prompt instruction, Qwen2.5-14B generated:
\begin{minipage}{\columnwidth}
\begin{small}
\begin{verbatim}
{"precond": {"obj": [0]},
 "exclude_precond": {"key-package": [0]},
 "delete_objects": [0], ...}
\end{verbatim}
\end{small}
\end{minipage}
This rule protects key packages from deletion. Gemma3-12B used the field inconsistently, and other gating dependencies, including switch panels, remain unprotected. We report this extension as a schema case study, not as an additional quantitative result.

\subsection{Stage~2 Budget Policy Analysis}
\label{app:budget}

Runtime recovery competes with rule-based relaxation for the same deadline. Let $h$ be time remaining until $T/2$ at the recovery decision:
\begin{equation}\begin{aligned}
 t_{\mathrm{call}}&=\min(5,h-6),\\
 t_{\mathrm{retry}}&=\min(0.15T,h_{\mathrm{after\ call}}-5).
\end{aligned}\label{eq:recovery-budget}\end{equation}
A call starts only if its allowance is at least 1\,s; retry starts only with positive allowance, leaving 5\,s for rough planning when recovery is feasible.
The legacy comparison also requires $h\geq11$\,s. It skips the recorded 30\,s-setting opportunities, while the revised policy can allocate a shorter call.
This is a resource safeguard, not evidence that the LLM chooses better objects than equal-count random recovery.

\subsection{Exploratory Stage~3 Prompt Variants}
\label{app:stage3_improvements}

The submitted prompt studies used unmatched samples and logging conditions; their single-seed rankings are not carried forward as controlled evidence.
The revision evaluates direct scoring, 32-object batching and goal/batch-related input, with bounded policies separated from the earlier unbounded configuration (\sref{sec:results_stage3}).
Fallback replay applies 0.1/0.3/0.5/0.7/0.9 to identical raw scores; the supplement retains every 50-map replay row and raw response coverage. Replays are not new online calls.
The later compact-hint study limits the requested output to eight cells and separately controls the returned information, model architecture and inference overhead.
These changes distinguish output omission, invalid scores, input-information limits and latency; none alone proves that the information is useful for planning.

\subsection{Baseline Comparison: Pure FD, PLOI, and Flax Variants}
\label{app:baselines}

This subsection compares four configurations that isolate unguided symbolic search, GNN scoring, and manual or generated rules.
Pure FD uses no guidance or rules. PLOI uses the GNN-guided \texttt{IncrementalPlanner}~\cite{silver2021planning} without relaxation or complementary rules.
Manual Flax and \method use the same GNN with the rule sets specified in Setup; the latter is distinct from \fullmethod.
The four-way SR and failure-capped timing comparison is consolidated in Table~\ref{tab:main}, with the eight-setting SR overview in Figure~\ref{fig:sr_heatmap}.

\myparagraph{Main observations}
Pure FD solves all instances in five of the eight settings and must therefore be treated as a competitive baseline rather than a lower bound. Its SR falls to 42.0\% on $12{\times}12$ Expert and 38.0\% on $15{\times}15$ Hard.
Because time is conditioned on success, a lower success-only mean does not by itself establish greater efficiency: a configuration that solves fewer, easier instances can have a lower mean.
Both rule configurations achieve higher SR than the rule-free configurations on $12{\times}12$ Expert and $15{\times}15$ Hard, while Pure FD is strongest on several easier settings.
These results support generated rules within Flax, with higher mean SR than the manual configuration across eight settings; the favorable rule content is not unique to an LLM generator (\sref{sec:model_comparison}).

\subsection{Qualitative Visualization: Planning Traces}
\label{app:vis}

The qualitative panels from the submitted manuscript are preserved as archived illustrations: (a) the full problem, (b) zero-shot LLM scores for diagnostic visualization, and (c) the retained set and plan from the original Stage~1 configuration. They are not new measurements or evidence for the revised tables.
The scores in column (b) are not used to produce column (c); the planner uses the trained GNN, matching \tref{tab:main}.
The left block covers the four smaller grids ($10{\times}10$ and $12{\times}12$ Medium);
the right block covers the four harder grids ($12{\times}12$ Hard/Expert and $15{\times}15$).

\begin{sidewaysfigure*}
  \centering
  \includegraphics[width=0.93\textheight, keepaspectratio]{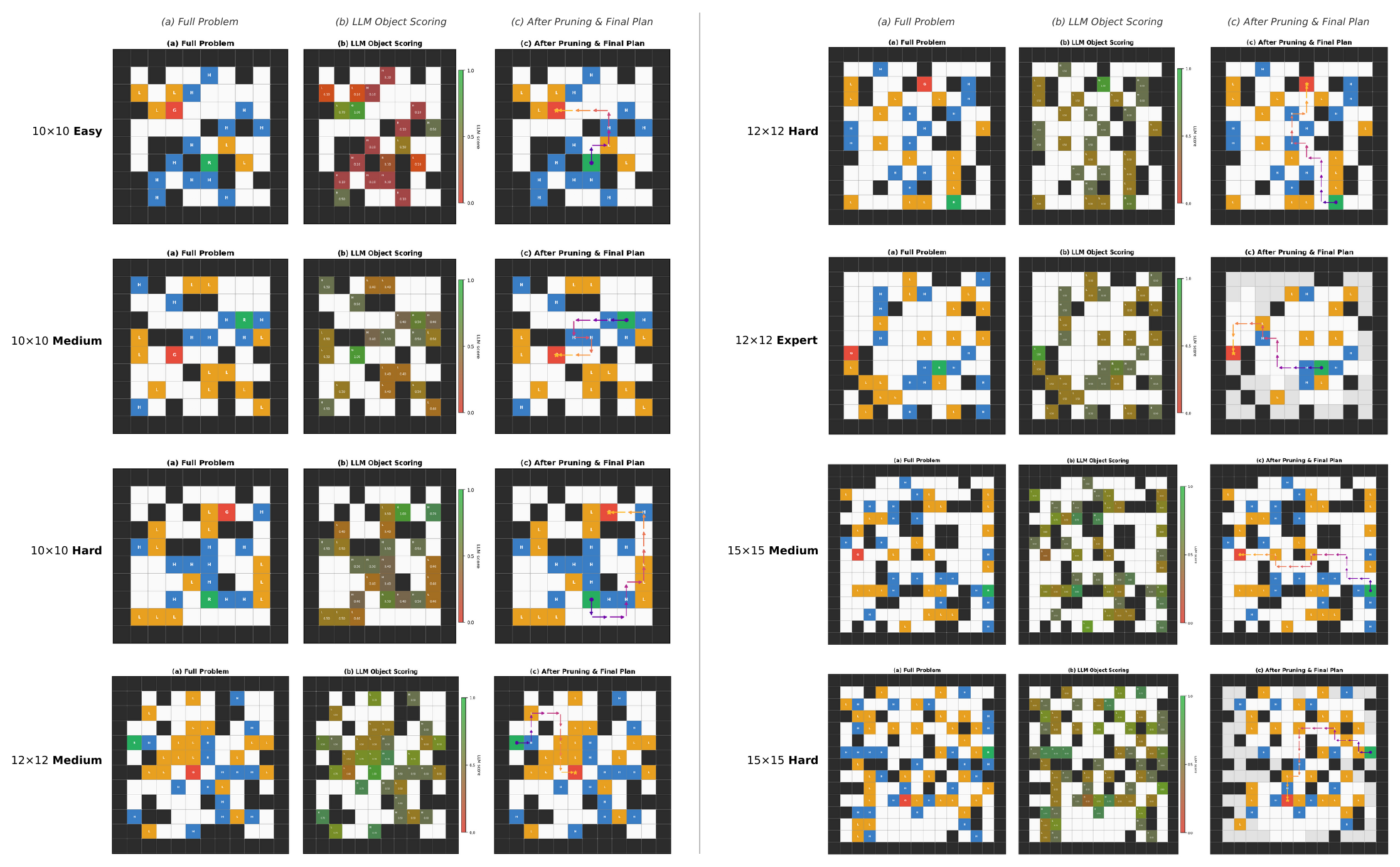}
  \caption{Archived qualitative MazeNamo traces from the submitted manuscript. Each row is one benchmark; columns show \textbf{(a)} the full problem, \textbf{(b)} zero-shot LLM scores for diagnostic display only, and \textbf{(c)} the retained object set and plan produced by the Stage~1 configuration using the trained GNN and generated rules.}
  \label{fig:vis_all}
\end{sidewaysfigure*}

\begin{figure*}[!t]
  \centering
  \includegraphics[width=0.95\textwidth]{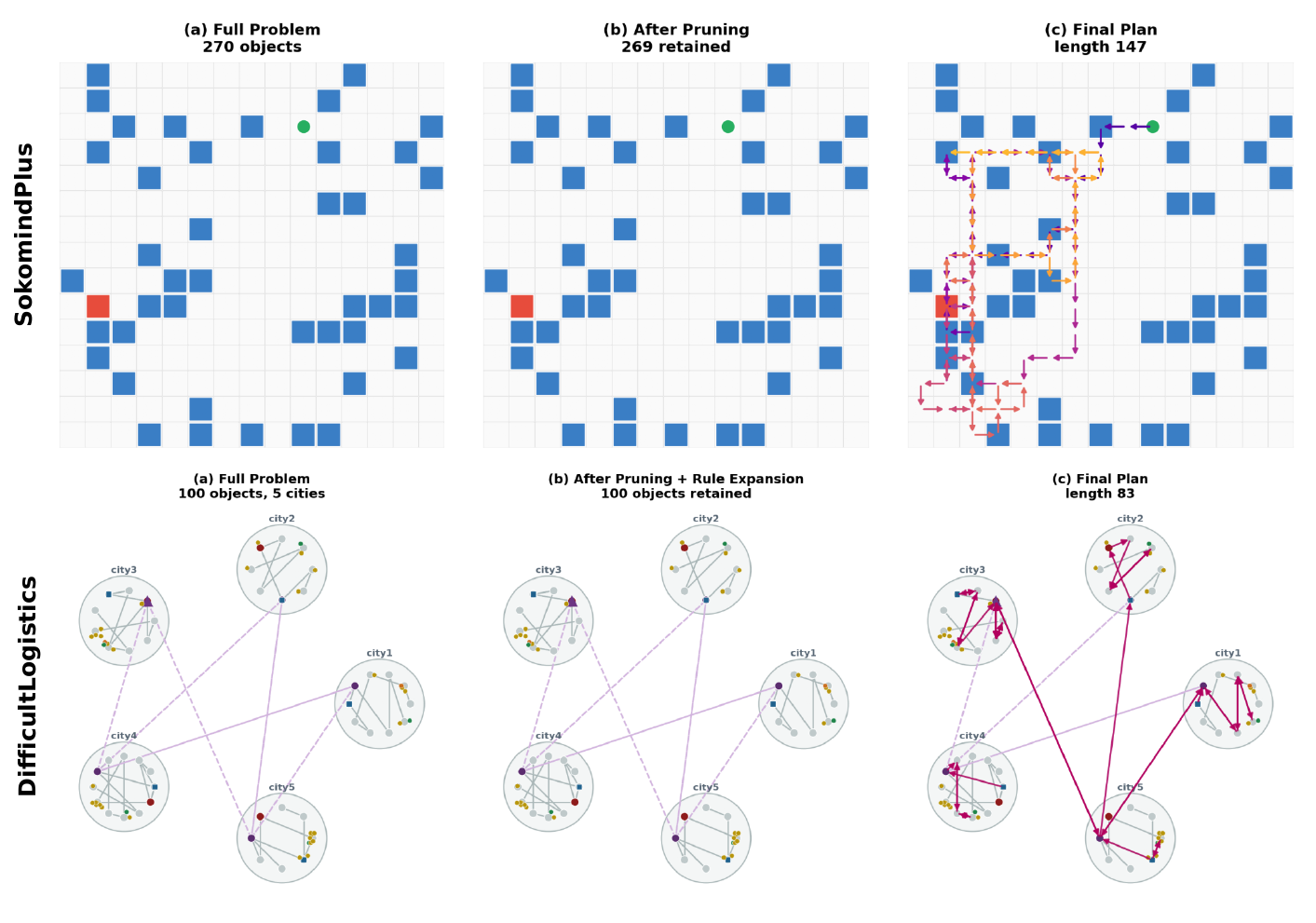}
  \caption{Archived qualitative traces from the submitted manuscript with stored Gemma3-12B rules and the no-guidance scorer; these are separate from the new domain-trained GNN results. \textbf{Top}: SokomindPlus; blue denotes boxes, red the goal box, and green the robot. \textbf{Bottom}: DifficultLogistics; locations are grouped by city, dashed edges denote air links, and column (c) traces the truck and airplane routes. The DifficultLogistics trace retains all 100 objects after the relaxed problem is found unsolvable.}
  \label{fig:vis_crossdomain}
\end{figure*}

\fref{fig:vis_crossdomain} shows one solved instance from each additional domain using generated rules and the no-guidance scorer. SokomindPlus is rendered as a grid; DifficultLogistics is rendered as a city-clustered graph. In the latter trace, the relaxed problem is unsolvable, so the corrected planner retains all 100 objects before computing the final plan.

\subsection{Stage~3: Zero-Shot Object Scoring}\label{sec:results_stage3}
The updated scoring results retain direct zero-shot scoring and add bounded batch/local-fact variants. All three use Qwen2.5 14B with temperature 0, seed 0 and a 16,384-token context.
Direct scoring permits one request of at most 5\,s within the global deadline. The bounded batch/local variants limit total scoring to $\min(5\,\mathrm{s},0.2h)$, where $h$ is remaining global time at scoring start; each batch has at most one omission-correction request. Each uses the same 50 problem IDs per setting; their inference policies differ and are stated explicitly.
\begin{table}[tbp]\centering\small
\caption{Scoring SR (\%) on 50 maps per setting. Direct/Batch/Local each use one inference setting; GNN is mean $\pm$ SD over ten seeds. Batch/Local also change inference allowances; full timing/coverage is in the supplement.}\label{tab:stage3}\label{tab:runtime0}
\setlength{\tabcolsep}{3pt}
\begin{tabular}{lrrrr}\toprule
Task & Direct & Batch & Local & GNN\\\midrule
12X & 10.0 & 18.0 & 34.0 & 74.6 $\pm$ 2.3\\
12H & 96.0 & 88.0 & 88.0 & 95.0 $\pm$ 2.4\\
15M & 78.0 & 76.0 & 76.0 & 93.8 $\pm$ 1.5\\
15H & 24.0 & 34.0 & 32.0 & 94.6 $\pm$ 3.0\\
\bottomrule\end{tabular}\end{table}

\myparagraph{Diagnosing the Stage~3 failure directly}\label{sec:stage3_diagnosis}
Raw score coverage, invalid values and per-call latency are retained separately. An incomplete dictionary triggers the policy fallback and is not a calibrated measure of uncertainty or object necessity.
The small-label 20/50 comparisons, their constant-feature controls and runtime masks are included in the supplement.

\myparagraph{Completed full-data hybrid study}\label{sec:runtime-final}
The compact-hint study chooses Qwen2.5 14B on development/validation evidence, then fixes temperature 0, seed 0, context 2,048, output cap 96 tokens and at most $\min(3,0.1h)$ seconds for one online request, where $h$ is remaining global time.
The prompt asks for at most eight cell hints. A valid response is not a dense score dictionary or 100\% object coverage.
Twenty full-data hybrid/constant checkpoints across ten seeds pass gradient, parameter-update, overfit and learning-curve gates.
Same-weight masking, within-state shuffled hints and separately trained constant-feature controls all pay the actual online call cost. The GNN-only reference makes no LLM call.

\begin{table}[tbp]\centering\small
\caption{Primary runtime cohort: 300 new maps, ten GNN seeds. SR is percent (mean $\pm$ SD), $t_c$ failure-capped seconds. Intervals are paired SR differences versus GNN in percentage points. Other cohorts and pairwise controls are in the supplement.}\label{tab:runtime-final}
\setlength{\tabcolsep}{3pt}
\begin{tabular}{lrrr}\toprule
Arm & SR (\%) & $t_c$ & 95\% CI (pp)\\\midrule
GNN & 89.7 $\pm$ 0.5 & 3.53 & ---\\
Hybrid & 89.7 $\pm$ 0.5 & 4.27 & [-1.17,+0.97]\\
Masked & 90.3 $\pm$ 0.5 & 4.29 & [-0.33,+1.57]\\
Shuffled & 90.0 $\pm$ 0.4 & 4.20 & [-0.93,+1.47]\\
Constant & 90.0 $\pm$ 0.4 & 4.05 & [-0.37,+1.07]\\
\bottomrule\end{tabular}\end{table}

On the 300-new-map cohort, hybrid SR is 89.67\% versus GNN 89.73\%, with paired interval $[-1.17,+0.97]$ percentage points and capped time 4.27 versus 3.53\,s.
Information controls do not establish an LLM-information advantage. The primary cohort contributes $300\times10\times5=15{,}000$ executions (maps $\times$ GNN seeds $\times$ arms). The study-wide total of 38,200 adds 3,200 development and 20,000 exposed-replication executions; it is not the primary-cohort sample size. Across these cohorts, 35,252 executions pass full-problem VAL. The four non-GNN arms account for 30,560 online observations, each with one recorded request.
The learned hybrid uses 195 successful labels and is not zero-shot. This runtime result uses a local model; GPT is evaluated only for offline rules.\footnote{Portions drafted or revised with OpenAI Codex~\cite{openai2026codex}; see Acknowledgments.}

\section{Discussion}
\label{sec:discussion}

\myparagraph{Evidence for Stage~1}
Stage~1 is the best-supported component of \method. Gemma3-12B reproduces the manual MazeNamo relaxation rule in the fixed configuration and generates complementary relations that yield higher mean SR across all eight settings.
The mean SR increases from 90.95\% to 93.93\%; on 12 Hard it increases from 87.2\% to 95.0\% (+7.8 percentage points; paired 95\% interval [+2.4,+15.0]).
The validator constrains syntax and predicate references; it cannot establish that an accepted rule preserves useful planning semantics. The performance variation across models and the DifficultLogistics gating-object case illustrate this remaining requirement.
GPT produces five active relaxation proposals in five first-call attempts and reaches the manual/action reference SR of 90.0\% on the 40-map cohort.
Matched-budget controls distinguish rule-content effects from generator choice: GPT C2/C4 selections coincide with the PDDL-action control, and a generic binary enumerator reproduces the fixed rules' complementary expansions on the 355 inspected states.

\myparagraph{Role of the symbolic baseline}
Pure FD reaches SR~100.0\% on five MazeNamo settings and on SokomindPlus. Its performance falls on the two hardest MazeNamo settings.
Manual and generated rules improve SR in those settings, but the result is a benchmark-dependent trade-off rather than a uniform advantage over unguided search.
Success-only mean time is conditioned on the solved set and should not be compared independently of SR. Failure-capped and common-success time make this distinction explicit; PLOI is strongest among the evaluated configurations on DifficultLogistics.

\myparagraph{Runtime LLM guidance}
Stage~2 does not improve overall SR across the four evaluated benchmarks. Its budget policy reserves time for relaxation; equal-count random recovery remains competitive.
The direct Stage~3 diagnostic shows that single-turn JSON generation omits most objects. The revision tests bounded batching, information selection and a separately learned hybrid rather than treating batching as future work.
The hybrid reaches 89.67\% versus GNN's 89.73\% on 300 newly sampled maps (paired interval $[-1.17,+0.97]$ percentage points), with higher failure-capped time.
Masking, shuffling and constant-feature controls do not establish an LLM-information gain in this configuration.

\myparagraph{Transfer and adaptation cost}
The additional-domain experiments compare rule sources under domain-trained scorers and five GNN seeds. Generated rules reach 86.8\% versus manual 64.4\% on SokomindPlus and 92.4\% versus 82.8\% on DifficultLogistics.
They show that the rule-generation procedure can produce usable configurations outside MazeNamo, including a structurally different logistics domain, but three domains are insufficient to establish broad domain generalization.
Stage~1 also retains the GNN trained on 195 successful MazeNamo labels from 200 attempts. Rule construction and the cost of domain-specific training data are therefore evaluated separately; machine-call cost does not measure human-labor savings.

\myparagraph{Limitations}
The fixed-rule experiments use repeated GNN seeds and matched problem sets. The hosted generation pilot instead has five sessions, 40 exposed development maps and one GNN checkpoint; its exploratory intervals are unadjusted.
The hosted model, provider, decoding and resource budgets change together, so this comparison does not isolate model size. The finite initial-state audit does not prove universal rule equivalence, generation reliability or completeness.
Incomplete historical cohorts remain excluded from complete-grid comparisons. A separate guarded-search intervention gives identical 88.33\% primary confirmation SR without new LLM information.
The PDDL benchmarks are deterministic and fully specified; output omission, invalid scores, rule errors and latency are evaluated, while observation noise and stochastic dynamics are not modeled.\footnote{Portions drafted or revised with OpenAI Codex~\cite{openai2026codex}; see Acknowledgments.}

\section{Conclusion}
\label{sec:conclusion}

We presented three LLM integrations for the Flax neuro-symbolic planner and evaluated their planning performance and computational costs separately.
The Stage~1 configuration replaces manually authored rules while retaining the trained GNN. It obtains a mean SR of 93.93\% across eight MazeNamo settings, compared with 90.95\% for manual rules, and improves the evaluated manual baselines in two additional domains.
The controlled rule-set ablations and matched-budget comparisons distinguish rule-content gains from generator choice.
GPT-6 Astra resolves the observed local relaxation-generation failures with five active first-call proposals and reaches the manual/action reference SR of 90.0\% in the component-isolated pilot.

Stage~2 reserves fallback time but does not improve overall SR. The learned runtime hybrid does not exceed the GNN on the primary new-map cohort and retains training-data and inference costs.
Taken together, the experiments establish schema-constrained LLM rule construction and measured planning gains from the evaluated rule configuration.
They do not yet support zero-shot LLM scoring as a performance-matched replacement for the trained object scorer.\footnote{Portions drafted or revised with OpenAI Codex~\cite{openai2026codex}; see Acknowledgments.}

\section*{Acknowledgments}
This research was  supported by research funds for newly appointed professors of Jeonbuk National University in 2025.
Also, this work builds directly on the Flax neuro-symbolic planner~\cite{du2026fast}.
The core three-step planning loop, the MazeNamo benchmark environments, and the GNN training infrastructure are all from the original Flax system.
We thank the authors for making their code and benchmark publicly available.

OpenAI Codex~\cite{openai2026codex} was used to draft and revise portions of the Abstract, Sections~I--VI and Supplement; OpenAI image-generation tools~\cite{openai2026image} generated Figure~\ref{fig:architecture}. The authors verified and take responsibility for the content.

\phantomsection
\bibliographystyle{IEEEtran}
\bibliography{refs}

\begin{IEEEbiography}[{\includegraphics[width=1in,height=1.25in,clip,keepaspectratio]{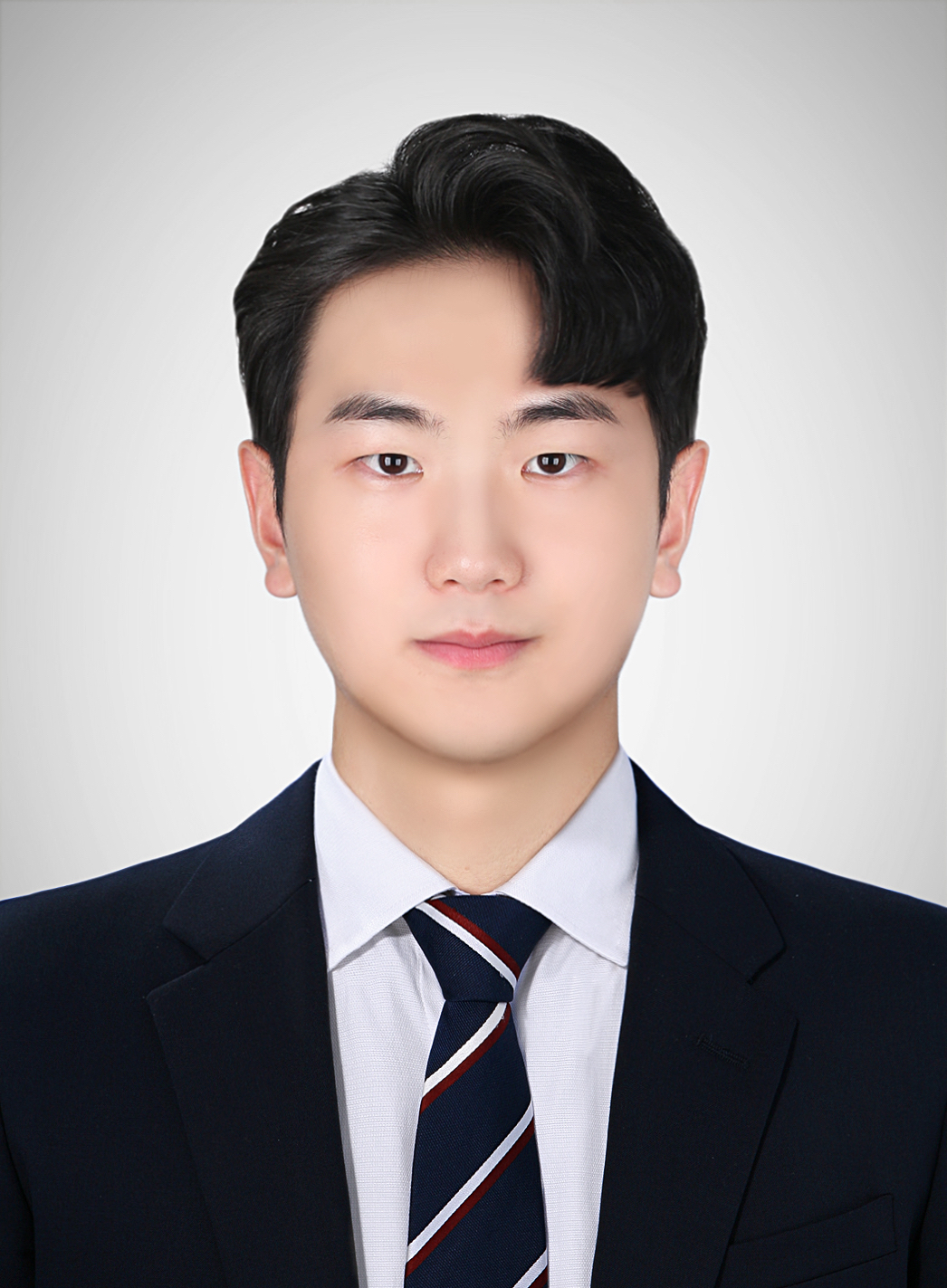}}]{Seongmin Kim}
received the B.S. degree in mechanical system engineering from Jeonbuk National University, Jeonju, Republic of Korea, in 2026. He is currently pursuing the M.S. degree with the Department of Physical-AI Convergence Engineering, Jeonbuk National University, Jeonju, Republic of Korea. His research interests include physical AI, manned-unmanned teaming (MUM-T), and autonomous systems.
\end{IEEEbiography}

\begin{IEEEbiography}[{\includegraphics[width=1in,height=1.25in,clip,keepaspectratio]{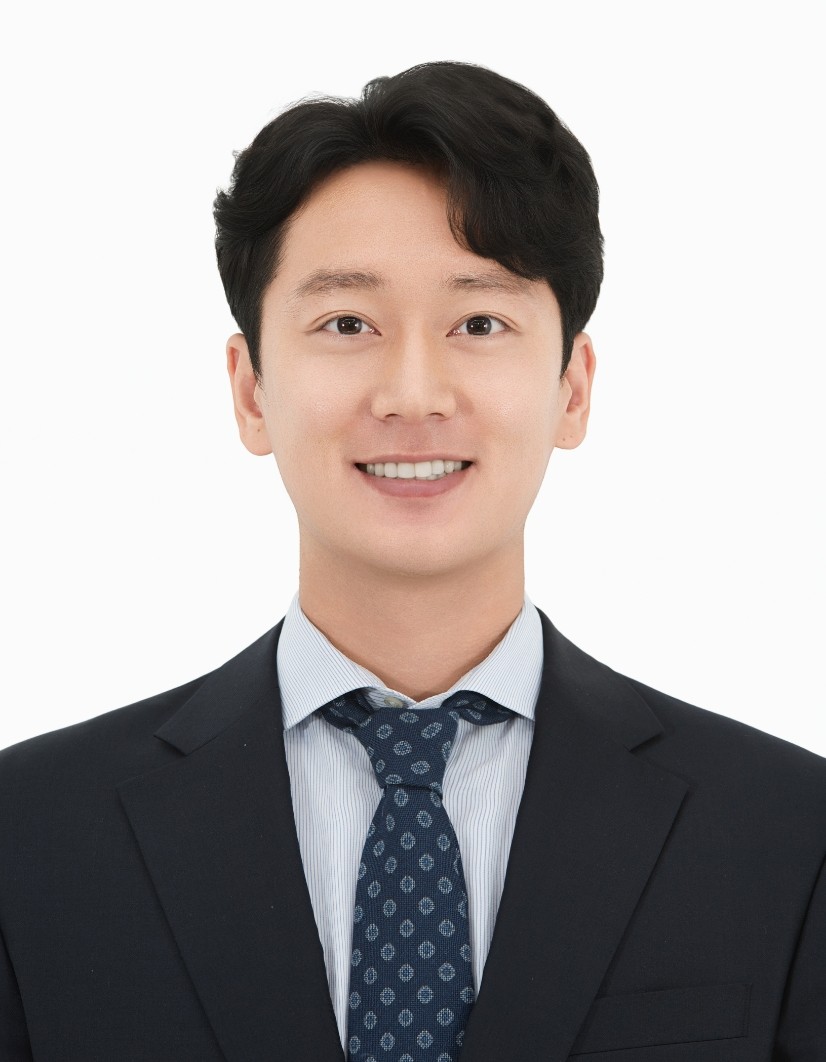}}]{Daegyu Lee}
received the B.S. degree in automotive engineering from Kookmin University, Seoul, Republic of Korea, in 2018, and the M.S. and Ph.D. degrees in future vehicle and electrical engineering from the Korea Advanced Institute of Science and Technology (KAIST), Daejeon, Republic of Korea, in 2020 and 2024, respectively. He was previously a Researcher with the Electronics and Telecommunications Research Institute (ETRI), Daejeon, Republic of Korea. Since October 2025, he has been an Assistant Professor with the Department of Advanced Defense Technology and Industry, Jeonbuk National University, Jeonju, Republic of Korea. His research interests include autonomous systems and localization, with a focus on spatial AI for environmental perception, learning-based navigation, and intelligent autonomous systems for defense applications.
\end{IEEEbiography}

\EOD

\end{document}